\documentclass[11pt,a4paper]{article}
\usepackage[hyperref]{acl2019}
\usepackage{times}
\usepackage{latexsym}
\usepackage{eurosym}

\usepackage{url}
\usepackage{caption}
\usepackage{graphicx}
\usepackage{multirow}
\usepackage{tabularx}
\usepackage{algorithm}
\usepackage{algorithmic}
\usepackage{amsmath}
\usepackage{arydshln}
\usepackage{color,soul}

\aclfinalcopy 
\def\aclpaperid{120} 

\title{Hierarchical Document Encoder for Parallel Corpus Mining}

\author{
Mandy Guo,
Yinfei Yang,
Keith Stevens,
Daniel Cer, 
Heming Ge,
\\ \rm\textbf{
Yun{-}Hsuan Sung,
Brian Strope,
Ray Kurzweil
} \AND
  {\rm Google AI}\\1600 Amphitheatre Parkway\\Mountain View, CA, USA\\
 \{xyguo, yinfeiy, kstevens, cer, hemingge, yhsung, raykurzweil\}@google.com}

\date{}

\begin{document}
\maketitle
\begin{abstract}
We explore using multilingual document embeddings for nearest neighbor mining of parallel data. Three document-level representations are investigated: (i) document embeddings generated by simply averaging multilingual sentence embeddings; (ii) a neural bag-of-words (BoW) document encoding model; (iii) a hierarchical multilingual document encoder (HiDE) that builds on our sentence-level model. The results show document embeddings derived from sentence-level averaging are surprisingly effective for clean datasets, but suggest models trained hierarchically at the document-level are more effective on noisy data. Analysis experiments demonstrate our hierarchical models are very robust to variations in the underlying sentence embedding quality. Using document embeddings trained with HiDE achieves state-of-the-art performance on United Nations (UN) parallel document mining, 94.9\% P@1\footnote{We use evaluation metrics precision at N, here P@1 means precision at 1} for en-fr and 97.3\% P@1 for en-es. 
\end{abstract}

\section{Introduction}

Obtaining a high-quality parallel training corpus is one of the most critical issues in machine translation. Previous work on parallel document mining using large distributed systems has proven effective~\cite{jakob2010,antonova2011}, but these systems are often heavily engineered and computationally intensive. Recent work on parallel data mining has focused on sentence-level embeddings~\cite{mandy2018,DBLP:journals/corr/abs-1811-01136,marginloss}. However, these sentence embedding methods have had limited success when applied to document-level mining tasks ~\cite{mandy2018}.
A recent study from~\newcite{marginloss} shows that document embeddings obtained from averaging sentence embeddings can achieve state-of-the-art performance in document retrieval on the United Nation~(UN) corpus. This simple averaging approach, however, heavily relies on high quality sentence embeddings and the cleanliness of documents in the application domain.


In our work, we explore using three variants of document-level embeddings for parallel document mining: (i) simple averaging of embeddings from a multilingual sentence embedding model~\cite{marginloss}; (ii) trained document-level embeddings based on document unigrams; (iii) a simple hierarchical document encoder (HiDE) trained on documents pairs using the output of our sentence-level model.

The results show document embeddings are able to achieve strong performance on parallel document mining.
On a test set mined from the web, all models achieve strong retrieval performance, the best being 91.4\% P@1 for en-fr and 81.8\% for en-es from the hierarchical document models.
On the United Nations (UN) document mining task~\cite{uncorpus}, our best model achieves 96.7\% P@1 for en-fr and 97.3\% P@1 for en-es, a 3\%+ absolute improvement over the prior state-of-the-art~\cite{mandy2018,jakob2010}.
We also evaluate on a noisier version of the UN task where we do not have the ground truth sentence alignments from the original corpus.
An off-the-shelf sentence splitter is used to split the document into sentences.\footnote{To introduce noise in sentence alignment, which is often seen in the real applications, in the parallel documents}
The results shows that the HiDE model is robust to the noisy sentence segmentations, while the averaging of sentence embeddings approach is more sensitive.
We further perform analysis on the robustness of our models based on different quality sentence-level embeddings, and show that the HiDE model performs well even when the underlying sentence-level model is relatively weak.

We summarize our contributions as follows:
\begin{itemize}
\item We introduce and explore different approaches for using document embeddings in parallel document mining. 
\item We adapt the previous work on hierarchical networks to introduce a simple hierarchical document encoder trained on document pairs for this task.
\item Empirical results show our best document embedding model leads to state-of-the-art results on the document-level bitext retrieval task on two different datasets. The proposed hierarchical models are very robust to variations in sentence splitting and the underlying sentence embedding quality.
\end{itemize}

\section{Related Work}
Parallel document mining has been extensively studied. One standard approach is to identify bitexts using metadata, such as document titles \cite{yang2002mining}, publication dates \cite{munteanu2005improving,munteanu2006extracting}, or document structure \cite{chen2000parallel,resnik2003web,shi2006dom}. However, the metadata related to the documents can often be sparse or unreliable \cite{jakob2010}. More recent research has focused on embedding-based approaches, where texts are mapped to an embedding space to calculate their similarity distance and determine whether they are parallel \cite{gregoire2017,hassan2018achieving,schwenk2018filtering}. \citet{mandy2018} has studied document-level mining from sentence embeddings using a hyperparameter tuned similarity function, but had limited success compared to the heavily engineered system proposed by~\newcite{jakob2010}. 

An extensive amount of work has also been done on learning document embeddings.~\citet{le2014,li2016,dai2015} explored Paragraph Vector with various lengths (sentence, paragraph, document) trained on next word/n-gram prediction given context sampled from the paragraph. The work from~\citet{roy2016,chen2017,wu2018} obtained document embeddings from word-level embeddings. More recent work has been focused on learning document embeddings through hierarchical training. The work from \citet{yang-etal-2016-hierarchical, miculicich-etal-2018-document} approached Document Classification and Neural Machine Translation using Hierarchical Attention Networks, and \citet{wang-etal-2017-exploiting-cross} proposed using a hierarchy of Recurrent Neural Networks (RNNs) to summarize the cross-sentence context. However, the amount of work applying document embeddings to the translation pair mining task has been limited.

\citet{marginloss} recently showed strong parallel document retrieval results using document embeddings obtained by averaging sentence embeddings. 
Our paper extends this work to explore different variants of document-level embeddings for parallel document mining, including using an end-to-end hierarchical encoder model.

\begin{figure}[!tbp]
  \centering
  \includegraphics[width=.35\textwidth]{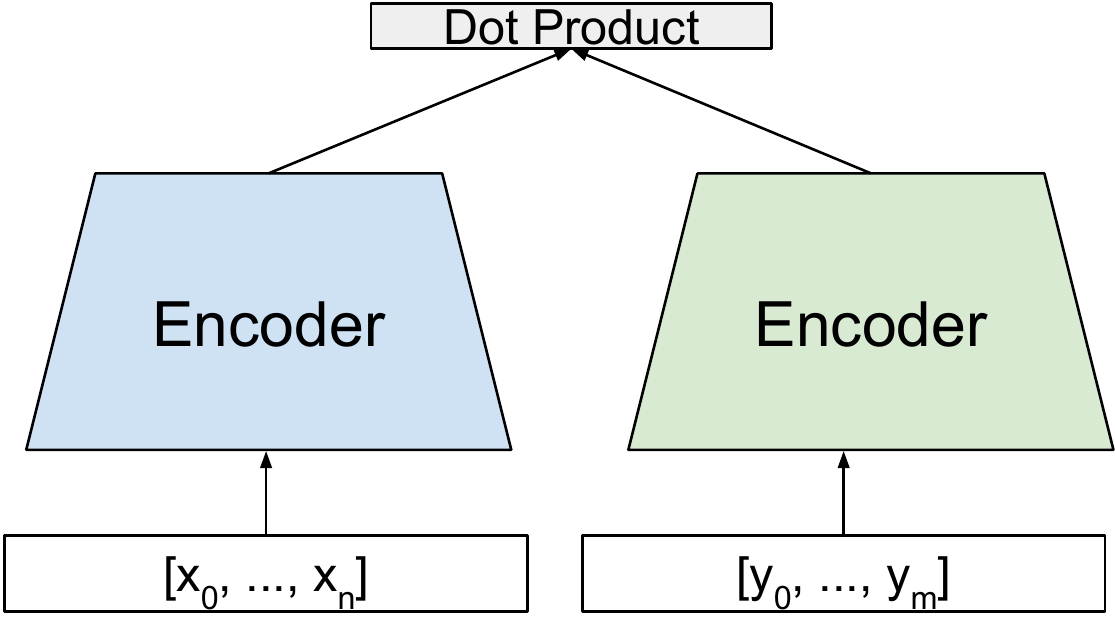}
  \caption{
    Dual encoder for parallel corpus mining, where $(x, y)$ represents translation pairs.
  }
  \label{fig:dual_encoder}
\end{figure} 

\begin{figure*}[!htbp]
  \centering
  \includegraphics[width=.95\textwidth]{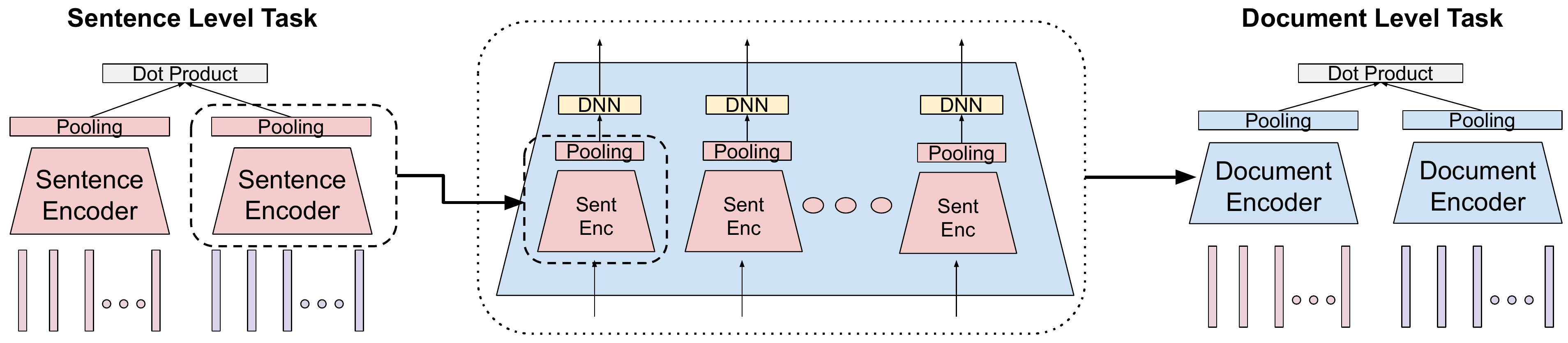}
  \caption{
  Illustration of the ${\text{DNN} \rightarrow \text{pooling}}$ version of the Hierarchical Document Encoder (HiDE). Each sentence is processed by our Transformer based encoding model with the final sentence-level embedding being produced by pooling across the last layer's positional heads. Document-level embeddings are composed by pooling across the sentence-level embeddings after each sentence embedding has been adapted by additional feed-forward layers.
  }
  \label{fig:hierarchy}
\end{figure*}

\section{Model}
This section introduces our document embedding models and training procedure.

\subsection{Translation Candidate Ranking Task using a Dual Encoder}

All models use the dual encoder architecture in Figure \ref{fig:dual_encoder}, allowing candidate translation pairs to be scored using an efficient dot-product operation. The embeddings that feed the dot-product are trained by modeling parallel corpus mining as a translation ranking task \cite{mandy2018}.
Given translation pair $(x, y)$, we learn to rank true translation $y$ over other candidates, $\mathcal{Y}$. We use batch negatives, with sentence $y_i$ of the pair $(x_i, y_i)$ serving as a random negative for all source $x_j$ in a training batch such that $j \neq i$. 
Following \newcite{DBLP:journals/corr/abs-1811-01136}, a shared multilingual encoder is used to map both $x$ and $y$ to their embedding space representations $x'$ and $y'$. Within a batch, all pairwise dot-products can be computed using a single matrix multiplication. We train using additive margin softmax \cite{marginloss}, subtracting a margin term $m$ from the dot-product scores for true translation pairs. For batch size $K$ and margin $m$, the log-likelihood loss function is given by Eq. \ref{eq:softmax}. 

\begin{equation}
\label{eq:softmax}
\mathcal{J} = -\frac{1}{K} {\sum_{i=1}^{K} \log \frac{e^{x'_i \cdot y_i^{\prime\top} - m}} { e^{x'_i \cdot y_i^{\prime\top} - m} + \sum_{k=1}^{K}e^{x'_{k,k\neq i} \cdot y_k^{\prime\top}}}}
\end{equation}

Models are trained with a bidirectional ranking objective \cite{marginloss}. Given source and target pair $(x, y)$, forward translation ranking, $\mathcal{J}_{forward}$, maximizes $p(y|x)$, while backward translation ranking, $\mathcal{J}_{backward}$, maximizes $p(x|y)$.\ Bidirectional loss $\mathcal{J}$ sums the two directional losses: 

\begin{equation}
\label{eq:softmax_prime}
\mathcal{J} = \mathcal{J}_{forward} + \mathcal{J}_{backward}
\end{equation}

\subsubsection{Sentence-Level Embeddings}
Sentence embeddings are produced by a Transformer model~\cite{DBLP:journals/corr/VaswaniSPUJGKP17} with pooling over the last block.\footnote{For pooling, we concatenate the combination of min, max and attentional pooling.} Semantically similar hard negatives are included to augment batch negatives ~\cite{mandy2018,mutty2018,marginloss}. We denote document embeddings derived from averaged sentence embeddings as \textbf{Sentence-Avg}.

\subsubsection{Bag-of-words Document Embeddings}

Our bag-of-words (BoW) document embeddings, \textbf{Document BoW}, are constructed by feeding document unigrams into a deep averaging network (DAN)~\cite{iyyer-EtAl:2015:ACL-IJCNLP} trained on the parallel document ranking task.\footnote{The model uses feed-forward hidden layers of size 320, 320, 500, and 500.}

\subsection{Hierarchical Document Encoder (HiDE)}

As illustrated in Figure \ref{fig:hierarchy}, our hierarchical model is also trained on the parallel document ranking task, but taking as input embeddings from our sentence-level model. For \textbf{HiDE$_{\text{DNN} \rightarrow \text{pooling}}$}, sentence embeddings are adapted to the document-level task by applying a feed-forward DNN to each sentence embedding. Average pooling aggregates the adapted sentence representations into the final fixed-length document embedding. We contrast performance with a variant of the model, \textbf{HiDE$_{\text{pooling} \rightarrow \text{DNN}}$}, that performs average pooling first followed by a feed-forward DNN to adapt the representation to document-level mining.

\begin{table*}[!ht]
    \small
    \centering
    \begin{tabular}{|l|p{0.85\textwidth}|}
        \hline
        \multicolumn{1}{|c|}{\bf Corpus} & \multicolumn{1}{l|}{\bf \hspace{4.1cm} Document Pairs} \\ \hline
        \multicolumn{2}{|c@{~~~~~~~~~~~~~~~~~~~~~~}|}{\rule{0pt}{8pt} \textit{English - French}} \\ \hline
        \rule{0pt}{12pt}
        \multirow{5}{*}{WebData}
        & (\boldmath$s_1$) Specs Toshiba Coverside FL not categorized (4407839940), ($s_2$) Search by brand, ($s_3$) Icecat: syndicator of product information via global Open catalog with more than 4578703 data-sheets \& 19844 brands -- Register (free) \\
        \rule{0pt}{12pt}
        & \boldmath($s_1$) Fiche produit Toshiba Coverside FL non class\'e (4407839940), ($s_2$) Partenaires en ligne, ($s_3$) Edit my products \\
        \cdashline{1-2}
        \rule{0pt}{12pt}
        \multirow{4}{*}{Clean UN} 
        & \boldmath($s_1$) 1 July 2011, ($s_2$) Original: English, ($s_3$) Tenth meeting, ($s_4$) Cartagena, Colombia, 17 - 21 October 2011, ($s_5$) Item 4 of the provisional agenda \\
        \rule{0pt}{12pt}
        & \boldmath($s_1$) 1er juillet 2011, ($s_2$) Original : anglais, ($s_3$) Dixi\`eme r\'eunion, ($s_4$) Cartagena (Colombie), 17-21 octobre 2011, ($s_5$) Point 4 de l'ordre du jour provisoire*\\ \cdashline{1-2}
        \rule{0pt}{12pt}
        \multirow{6}{*}{Noisy UN} 
        & \boldmath($s_1$) 6--7 May 1999 Non-governmental organizations New York, 14 to 18 December 1998 Corrigendum 1., ($s_2$) Paragraph 1, draft decision I, under ``Special consultative status'' 2., ($s_3$) Paragraph 48 Add Japan to the list of States Members of the United Nations represented by observers.\\ 
        \rule{0pt}{12pt}
        & \boldmath($s_1$) 6 et 7 mai 1999 Organisations non gouvernementales New York, 14-18 d\'ecembre 1998 Rectificatif Paragraphe 1, projet de d\'ecision I, sous la rubrique ``Statut consultatif sp\'ecial'' Paragraphe 48 Ajouter le Japon \`a la liste des \'Etats Membres de l'Organisation des Nations Unies repr\'esent\'es par des observateurs. \\ \hline
        \multicolumn{2}{|c@{~~~~~~~~~~~~~~~~~~~~~~}|}{\rule{0pt}{8pt} \textit{English - Spanish}} \\ \hline
        \rule{0pt}{12pt}
        \multirow{4}{*}{WebData}
        & \boldmath($s_1$) Alcudia travel Guide \& Map - android apps on Google play, ($s_2$) Travel \& Local, ($s_3$) Alcudia travel Guide \& Map, ($s_4$) Maps, GPS Navigation Travel \& Local, ($s_5$) Offers in-app purchases" \\
        \rule{0pt}{12pt}
        & \boldmath($s_1$) Beirut Travel Guide \& map - aplicaciones Android en Google play, ($s_2$) Todav\'ia m\'as '', ($s_3$) Selección de los editores, ($s_4$) Libros de texto, ($s_5$) Comprar tarjeta de regalo\\
        \cdashline{1-2}
        \rule{0pt}{12pt}
        \multirow{7}{*}{Clean UN} 
        & \boldmath($s_1$) [Original: English], ($s_2$) Monthly report to the United Nations on the operations of the Kosovo Force, ($s_3$) 1. Over the reporting period (1-28 February 2003) there were just over 26,600 troops of the Kosovo Force (KFOR) in theatre. \\
        \rule{0pt}{12pt}
        & \boldmath($s_1$) [Original: ingl\'es], ($s_2$) Informe mensual de las Naciones Unidas sobre las operaciones de la Fuerza Internacional de Seguridad en Kosovo, ($s_3$) En el per\'iodo sobre el que se informa (1° a 28 de febrero 2003) hab\'ia en el teatro de operaciones algo m\'as de 26.600 efectivos de la Fuerza Internacional de Seguridad en Kosovo (KFOR).\\
        \cdashline{1-2}
        \rule{0pt}{12pt}
        \multirow{7}{*}{Noisy UN} 
        & \boldmath($s_1$) (Original: English) Monthly report to the United Nations on the operations of the Kosovo Force 1., ($s_2$) Over the reporting period (1-28 February 2003) there were just over 26,600 troops of the Kosovo Force (KFOR) in theatre.\\ 
        \rule{0pt}{12pt}
        & \boldmath($s_1$) (Original: ingl\'es) Informe mensual de las Naciones Unidas sobre las operaciones de la Fuerza Internacional de Seguridad en Kosovo En el per\'iodo sobre el que se informa (1° a 28 de febrero 2003) hab\'ia en el teatro de operaciones algo m\'as de 26.600 efectivos de la Fuerza Internacional de Seguridad en Kosovo (KFOR). \\ \hline
    \end{tabular}
    \caption{Example document snippets from the WebData, original UN corpus, UN corpus with noisy sentence segmentation. We only show the starting sentences for each document, the original documents can go very long. Symbol ($s_n$) means sentence $n$ in the document to show sentence segmentation.}
    \label{tab:data}
\end{table*}

\section{Experiments}
This section describes our training data, model configurations, and retrieval results for our embedding models: Sentence-Avg, Document BoW, HiDE$_{\text{DNN} \rightarrow \text{pooling}}$, and HiDE$_{\text{pooling} \rightarrow \text{DNN}}$. 

\subsection{Data}
We focus on two language pairs: English-French (en-fr) and English-Spanish (en-es).
Two corpora are used for training and evaluation.

The first corpus is obtained from web (\textbf{WebData}) using a parallel document mining system and automatic sentence alignments, both following an approach similar to ~\newcite{jakob2010}.
Parallel documents number 13M for en-fr and 6M for en-es, with 400M sentence pairs for each language pair.
We split this corpus into training (80\%),  development (10\%), and test set (10\%).

We also evaluate the trained models on a second corpus, the United Nations (\textbf{UN}) Parallel Corpus~\cite{uncorpus}, as an out-of-domain test set.
The UN corpus contains a fully aligned subcorpus of $\sim$86k document pairs for the six official UN languages.\footnote{Arabic, Chinese, English, French, Russian, and Spanish.} As this corpus is small, it is only used for evaluation.

The sentence segmentation in the fully aligned subcorpus is particularly good due to the process used to construct the dataset. While automatic sentence splitting is performed using the Eserix spltter, documents are only included in the fully aligned subcorpus if sentences are consistently aligned across all six languages. This implicitly filters documents with noisy sentence segmentations. Exceptions are errors in the sentence segmentation that are systematically replicated across the documents in all six languages. 

We create a noisier version of the UN dataset that makes use of an robust off-the-shelf sentence splitter, but which necessarily introduces noise compare to sentences that were split by consensus across all six languages within the original UN dataset. Models are evaluated on this noisy UN corpus, as any real application of our models will almost certainly need to contend with noisy automatic sentence splits.

Table \ref{tab:data} shows examples from each dataset. 
The WebData dataset is very noisy and contains a large amount of template-like queries from web. In this dataset, sentence alignments can be also very noisy, and sometimes sentences are not direct translations of each other.
The original UN is translated sentence by sentence by human annotators, so it is perfectly aligned at the sentence-level with ground truth translations.
The noisy UN, however, could have incorrect sentence-level mappings, but these could still be correct translations on the document-level.
The sentence splitter used to generate the noisy UN dataset could also perform differently in different languages for the parallel content, resulting in mismatches at the sentence-level.
As seen in the Noisy UN examples shown in Table \ref{tab:data}, the English text is split into 3 sentences, while the corresponding French or Spanish texts are only split into 1 sentence.

\begin{table*}
\centering
    \begin{tabular}{l | r r r || r r r } 
        \hline
        \multirow{2}{*}{Document Embedding} & \multicolumn{3}{c||}{en-fr (1M) } & \multicolumn{3}{c}{en-es (0.6M)} \\
        \cline{2-7}
        &  P@1 & P@3 & P@10  &  P@1 & P@3 & P@10 \\ 
        \hline
        HiDE$_{\text{DNN} \rightarrow \text{pooling}}$     & \textbf{91.40} & \textbf{94.13} & \textbf{95.67} & \textbf{81.83} & \textbf{87.85} & \textbf{91.45} \\
        HiDE$_{\text{pooling} \rightarrow \text{DNN}}$    & 90.63 & 93.50 & 95.11 & 78.84 & 85.04 & 88.88 \\ \hline
        Document BoW                 & 83.83 & 90.47 & 94.18 & 78.09 & 85.04 & 91.03 \\
        Sentence-Avg                 & 78.07 & 83.53 & 87.06 & 67.49 & 74.22 & 79.01 \\
        \hline
    \end{tabular}
\caption{Precision at N (P@N) of target document retrieval on the WebData test set. 
Models attempt to select the true translation target for a source document from the entire corpus (1 million parallel documents for en-fr, and 0.6 million for en-es).
}
\label{tab:eval_overlay}    
\end{table*}

\subsection{Configuration}
Our sentence-level encoder follows a similar setup as~\newcite{marginloss}. The sentence encoder has a shared 200k token multilingual vocabulary with 10K OOV buckets. Vocabulary items and OOV buckets map to 320 dim.\ word embeddings. For each token, we also extract character n-grams ($n=[3, 6]$) hashed to 200k buckets mapped to $320$ dim.\ character embeddings. Word and character n-gram representations are summed together to produce the final input token representation. Updates to the word and character embeddings are scaled by a gradient multiplier of 25~\cite{mutty2018}. The encoder uses 3 transformer blocks with hidden size of $512$, filter size of $2048$, and $8$ attention heads. Additive margin softmax uses $m=0.3$.
We train for 40M steps for both language pairs using an SGD optimizer with batch size K=100 and learning rate $0.003$. 

During document-level training, sentence embeddings are fixed due to the computational cost of dynamically encoding all of the sentences in a document. Sentence embeddings are adapted using a four-layer DNN model with residual connections and hidden sizes $320$, $320$, $500$, and $500$. The first three layers use ReLU activations with the final layer using Tanh. Document embeddings are trained with an SGD optimizer, batch size $K=200$, learning rate $0.0001$, and additive margin softmax $m=0.5$ for en-fr, and $m=0.6$ for en-es. We train for 5M steps for en-fr and 2M steps for en-es. Light hyperparameter tuning uses our development set from WebData.

\subsection{Mining Translations and Evaluation}
Translation candidates are mined with approximate nearest neighbor (ANN)~\cite{ann} search over our multilingual embeddings~\cite{mandy2018,DBLP:journals/corr/abs-1811-01136}.\footnote{Prior work only used ANN over sentence embeddings.} The evaluation metric is precision at N (P@N), which evaluates if the true translation is in the top N candidates returned by the model.

\subsubsection{Results on WebData Test Set}

Table \ref{tab:eval_overlay} presents document embedding P@N retrieval performance using our WebData test set, for N = $1$, $3$, $10$. The evaluation uses 1M candidate documents for en-fr and 0.6M for en-es. We obtain the best performance from our hierarchical models, HiDE$_*$. Adapting the sentence embeddings prior to pooling, HiDE$_{\text{DNN} \rightarrow \text{pooling}}$ performs better than attempting to adapt the representation after pooling, HiDE$_{\text{pooling} \rightarrow \text{DNN}}$. Document BoW embeddings outperform Sentence-Avg, showing training a simple model for document-level representations (DAN) outperforms pooling of sentence embeddings from a complex model (Transformer).

\subsubsection{Results on UN Corpus}

\begin{table}
\centering
    \begin{tabular}{l|r||r} 
        \hline
        Model & en-fr & en-es \\ 
        \hline
        \multicolumn{3}{c}{UN Corpus Sentence Segmentation} \\
        \hline
        HiDE$_{\text{DNN} \rightarrow \text{pooling}}$ & 96.6 & \textbf{97.3} \\
        HiDE$_{\text{pooling} \rightarrow \text{DNN}}$ & 96.5 & 96.1 \\
        Sentence-Avg & \textbf{96.7}  & \textbf{97.3} \\
        \hline
        \multicolumn{3}{c}{Noisy Sentence Segmentation} \\
        \hline
        HiDE$_{\text{DNN} \rightarrow \text{pooling}}$ & \textbf{94.9} & \textbf{96.0} \\
        HiDE$_{\text{pooling} \rightarrow \text{DNN}}$ & 91.0 & 94.4 \\
        Sentence-Avg & 86.8  & 95.7 \\
        \hline
        \multicolumn{3}{c}{No sentence splitting } \\
        \hline
        Document BoW & 74.3 & 71.9 \\
        \hline
        \multicolumn{3}{c}{\emph{Prior work}} \\
        \hline
        \newcite{jakob2010} & 93.4  & 94.4 \\
        \newcite{mandy2018} & 89.0  & 90.4 \\
        \hline
    \end{tabular}
\caption{Document matching on the UN corpus evaluated using P@1. For methods that require sentence splitting, we report results using both the UN sentence annotations and 
an off-the-shelf sentence splitter. }
\label{tab:un_combined}    
\end{table}

Table \ref{tab:un_combined} shows document matching P@1 for our models on both the original UN dataset sentence segmentation and on the noisier sentence segmentation. P@1 is evaluated using all of the UN documents in a target language as translation candidates. The prior state-of-the-art is~\newcite{jakob2010}.\footnote{\newcite{jakob2010} was applied to the UN dataset by \newcite{mandy2018}.} Using both the official and noisy sentence segmentations, HiDE$_{\text{DNN} \rightarrow \text{pooling}}$ outperforms~\newcite{jakob2010}, a heavily engineered system that incorporates both MT and monolingual duplicated document detection. 

\newcite{mandy2018} uses sentence-to-sentence alignments to heuristically identify document pairs. Alignments were computed using sentence embeddings generated over the UN corpus annotated sentence splits. With corpus annotated splits,  Sentence-Avg performs better than \newcite{mandy2018}. Furthermore, even with noisy sentence splits HiDE$_*$ outperforms \newcite{mandy2018}. 

The performance of all our document embeddings methods that build on sentence-level representations is remarkably strong when we use the sentence boundaries annotated in the UN corpus. Surprisingly, Sentence-Avg performed poorly on the WebData test data but is very competitive with both variants of HiDE when using the original UN corpus sentence splits.\footnote{We use similar sentence-level encoder setup as \citet{marginloss}, we are able to obtain matching results on the original UN corpus} However, on the UN data with noisy sentence splits, HiDE$_*$ once again significantly outperforms Sentence-Avg. Averaging sentence embeddings appears to be a strong baseline for clean datasets, but the hierarchical model helps when composing document embeddings from noisier input representations.\footnote{We note that in practice parallel document mining will tend to operate over noisy datasets.} Similar to the WebData test set, on the noisy UN data, HiDE$_{\text{DNN} \rightarrow \text{pooling}}$ outperforms HiDE$_{\text{pooling} \rightarrow \text{DNN}}$. We note that while Document BoW performed well on the in-domain test set, it performs poorly on the UN data. Preliminary analysis suggests this is due in part to differences in length between the WebData and UN documents.

\begin{figure}[!]
  \centering
  \includegraphics[width=.48\textwidth]{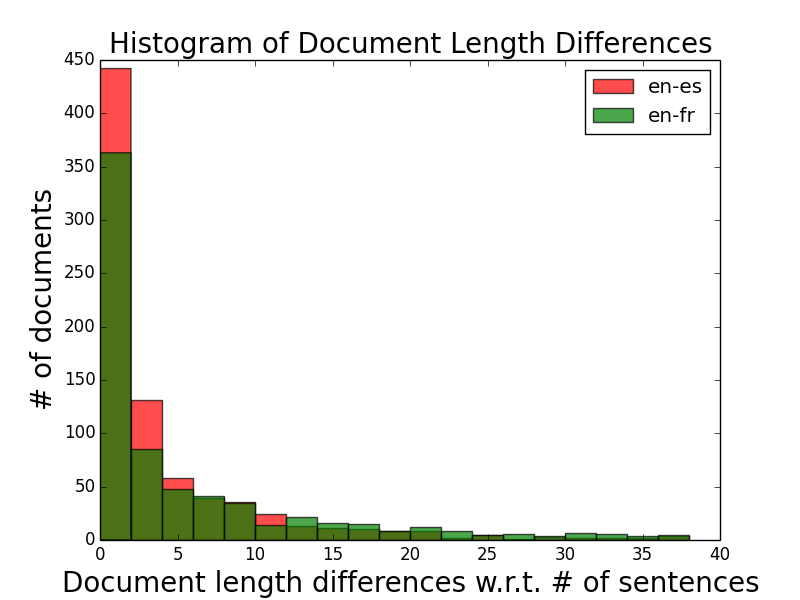}
  \caption{
    Histogram of document length differences w.r.t. \# of sentences in each parallel document pair.
  }
  \label{fig:hist_diff}
\end{figure}

We also observe that the performance of Sentence-Avg model dropped significantly in en-fr when transitioning from the Clean UN to the Noisy UN, but in en-es, the performance drop is much less.
We compute the histogram of the document length differences in each document pair w.r.t. the \# of sentences in each document on the noisy UN corpus.
As shown in figure \ref{fig:hist_diff}, the en-es dataset indeed has better agreement on the sentence split comparing with en-fr,
which indicates the Sentence-Avg model is sensitive to the sentence segmentation quality of the parallel document pairs.

\begin{table*}[!ht]
    \small
    \centering
    \begin{tabular}{|l|p{0.85\textwidth}|}
        \hline
        \multicolumn{2}{|c@{~~~~~~~~~~~~~~~~~~~~~~}|}{\rule{0pt}{8pt} \textit{Example 1}} \\ \hline
        \rule{0pt}{12pt}
        \multirow{2}{*}{Source}
        & Audio-technica mb 3k b-stock - Thomann ireland, Dynamic Microphones finder, 40 \euro -- 60, 60 \euro -- 100, 100 \euro -- 120, 120 \euro -- 160, 160 \euro -- 200, 200 \euro -- 280, 280 \euro -- 460, in stock items\\
        \cdashline{1-2}
        \rule{0pt}{12pt}
        \multirow{3}{*}{Expected Result} 
        & Beyerdynamic tg-x58 b-stock - Thomann espa\~na, Micr\'ofonos din\'amicos de voz encontrar ..., Gama de precios, 40 \euro -- 60, 60 \euro -- 100, 100 \euro -- 120, 120 \euro -- 160, 160 \euro -- 200, 200 \euro -- 280, 280 \euro -- 460, Reajustar todos los filtros\\
        \cdashline{1-2}
        \rule{0pt}{12pt}
        \multirow{3}{*}{Actual Result} 
        & Audio-technica atm63 u - Thomann espa\~na, Micr\'ofonos din\'amicos de voz encontrar ..., Gama de precios, 40 \euro -- 60, 60 \euro -- 100, 100 \euro -- 120, 120 \euro -- 160, 160 \euro -- 200, 200 \euro -- 280, 280 \euro -- 400, Reajustar todos los filtros \\ \hline
        \multicolumn{2}{|c@{~~~~~~~~~~~~~~~~~~~~~~}|}{\rule{0pt}{8pt} \textit{Example 2}} \\ \hline
        \rule{0pt}{12pt}
        \multirow{3}{*}{Source}
        & Casual man suit photo - android apps on google play, Casual man suit photo, Casual shirt Photo suit is photography application to make your face in nice fashionable man suit., This is so easy and free to make your photo into nice looking suit without any hard work and it's all free.\\
        \cdashline{1-2}
        \rule{0pt}{12pt}
        \multirow{3}{*}{Expected Result} 
        & Casual fotos - aplicaciones de android en Google play, Todav\'ia m\'as '', Selecci\'on de nuestros expertos, Libros de texto, Comprar tarjeta regalo, Mi lista de deseos, Mi actividad de Play, Gu\'ia para padres, Arte y Dise\~no, Bibliotecas y demos, Casa y hogar\\
        \cdashline{1-2}
        \rule{0pt}{12pt}
        \multirow{3}{*}{Actual Result} 
        & Traje de la foto de la camisa formal de los hombre - aplicaciones de android en Google play, Todav\'ia m\'as '', Selecci\'on de nuestros expertos, Libros de texto, Comprar tarjeta regalo, Mi lista de deseos, Mi actividad de Play, Gu\'ia para padres, Arte y Dise\~no, Bibliotecas y demos, Casa y hogar \\ \hline
    \end{tabular}
    \caption{Example document snippets of source, expected result, and actual result retrieved by HiDE$_{\text{DNN} \rightarrow \text{pooling}}$ model on the en-es development sets.}
    \label{tab:errors}
\end{table*}

\section{Analysis}

In this section, we first analyze the errors produced by the document embedding models.
We then explore how the performance of sentence-level models affect the performance of document-level models that incorporate sentence-embeddings.

\subsection{Errors}
We first look at the false positive examples retrieved by HiDE$_{\text{DNN} \rightarrow \text{pooling}}$ model on en-es WebDoc development set.
We observe that the actual error results often have similar sentence structure and meaning comparing to the expected result.

Table \ref{tab:errors} list two typical example snippets for HiDE$_{\text{DNN} \rightarrow \text{pooling}}$.
In the first example, our model matches the translation of "Audio-technica" to "Audio-technica" instead of "Beyerdynamic".
We observe that in multiple cases, HiDE model is able to retrieve a more accurate translation pair than the labeled expected result. 
As shown in Table \ref{tab:data}, the WebData automatically mined from the web is noisy and may contains non-translation pairs.
This results indicates the proposed model is robust to the training data noise.
The second example shows another typical error where the documents are template-like. 
The actual results retrieved by HiDE$_{\text{DNN} \rightarrow \text{pooling}}$ still largely match the expected text.

We also look at the actual results retrieved from Sentence-Avg model.
The Sentence-Avg model also suffers from the template-like documents (e.g. Example 2 in Table \ref{tab:errors}) similar to the HiDE$_{\text{DNN} \rightarrow \text{pooling}}$ model.
Other than that, though some correctly translated words can be found, the retrieved error documents differ much more in sentence structure and meaning from the expected results. For example, the expected and actual results can both be documents about the same subject, but from entirely different perspectives. We also found that some of the WebData target documents are in English instead of Spanish. In these cases, the Sentence-Avg model is more likely to retrieve a document in the same language as the source document instead of retrieving a translated document.

\begin{table*}[!htb]
    \centering
    \resizebox{\textwidth}{!}{\begin{tabular}{c | c | c c || c c }
        \hline
        \multirow{2}{*}{Languages} & \multirow{2}{*}{P@1 at Sentence Level} &\multicolumn{2}{c||}{P@1 on WebDoc test} & \multicolumn{2}{c}{P@1 on Noisy UN} \\
        \cline{3-6}
        & &  HiDE$_{\text{DNN} \rightarrow \text{pooling}}$ & Sentence-Avg &  HiDE$_{\text{DNN} \rightarrow \text{pooling}}$ & Sentence-Avg\\ \hline
        \multirow{4}{*}{en-fr}
        & 48.9 & 66.6 & 0.6 & 70.3 & 4.4 \\
        & 66.9 & 89.2 & 54.3 & 92.6 & 63.9 \\
        & 81.3 & 90.5 & 72.9 & 92.1 & 76.9 \\
        & 86.1 & 91.3 & 78.1 & 94.9 & 86.9 \\
        \hline
        \multirow{4}{*}{en-es}
        & 54.9 & 59.0 & 1.2 & 81.3 & 4.7 \\
        & 67.0 & 79.1 & 54.2 & 93.2 & 82.9 \\
        & 80.6 & 79.8 & 60.1 & 91.2 & 88.9 \\
        & 89.0 & 81.9 & 67.4 & 96.0 & 95.7 \\ \hline
    \end{tabular}}
    \caption{P@1 of target document retrieval on WebData test set and noisy UN corpus for HiDE$_{\text{DNN} \rightarrow \text{pooling}}$ and Sentence-Avg models with different sentence-level P@1 performance . The sentence-level peroformance is measured on the sentence-level UN retrieval task from the entire corpus (11.3 million sentence candidates).}
    \label{tab:worse_models}
\end{table*}

\subsection{HiDE performance on Coarse Sentence-level Models}
We further explore how the performance of sentence-level models affect the performance of document-level models that incorporate sentence-embeddings. We use different encoder configurations to produce sentence embeddings of varying quality as expressed by P@1 results for sentence-level retrieval on the UN dataset.\footnote{Model sentence-level model performance was varied by generating models with hyperparameters selected to degrade performance (e.g., fewer training sets, no margin softmax).}
Table \ref{tab:worse_models} shows the P@1 of target document retrieval on both the WebData test set and the noisy UN corpus for HiDE$_{\text{DNN} \rightarrow \text{pooling}}$ and Sentence-Avg. While sentence encoding quality does impact document-level performance, the HiDE model is surprisingly robust once the sentence encoder reaches around 66\% P@1, whereas the Sentence-Avg model requires much higher quality sentence-level embeddings (around 85\% for en-fr, and 80\% for en-es). The robustness of HiDE model provides a means for obtaining high-quality document embeddings without high-quality sentence embeddings, and thus provides the option to trade-off sentence-level embedding quality for speed and memory performance.

\section{Conclusion}
In this paper, we explore parallel document mining using several document embedding methods. Mining using document embeddings achieves a new state-of-the-art perfomance on the UN parallel document mining task (en-fr, en-es). Document embeddings computed by simply averaging sentence embeddings provide a very strong baseline for clean datasets, while hierarchical embedding models perform best on noisier data. Finally, we show document embeddings based on aggregations of sentence embeddings are surprisingly robust to variations in sentence embedding quality, particularly for our hierarchical models.

\section*{Acknowledgements}
We are grateful to the anonymous reviewers and our teammates in Deacartes and Google Translate for their valuable discussions, especially Chris Tar, Gustavo Adolfo Hernandez Abrego, and Wolfgang Macherey.

\bibliographystyle{acl_natbib}
\bibliography{acl2019}

\end{document}